\documentclass[11pt,a4paper]{article}
\usepackage[hyperref]{emnlp2020}
\usepackage{times}
\usepackage{latexsym}

\usepackage{url}
\usepackage{latexsym}
\usepackage{soul}
\usepackage{graphicx}
\usepackage{amsmath}
\usepackage{booktabs}
\usepackage{tabularx}
\usepackage{array,multirow}
\usepackage{subcaption}
\usepackage{float}
\usepackage{makecell}
\usepackage{framed}
\usepackage{enumitem}
\usepackage{titlesec} 
\usepackage{microtype}
\usepackage{siunitx}
\usepackage{xcolor}
\usepackage{multicol}
\usepackage{background}

\usepackage{microtype}

\setlist{nosep,topsep=3pt}

\definecolor{lexcolor}{rgb}{.1, .1, .4}
\newcommand{\lex}[1]{{\textit{#1}}}
\newcommand{\texttotext}[2]{\lex{#1} $\rightarrow$ \lex{#2}}

\newcolumntype{C}{>{\centering\arraybackslash}X}

\definecolor{hlcolor}{rgb}{.9, 1, .4}
\sethlcolor{hlcolor}

\colorlet{shadecolor}{hlcolor}
\newcommand{\hide}[1]{}

\newcommand{\blap}[1]{\smash[t]{\begin{tabular}[t]{@{}c@{}}#1\end{tabular}}}

\aclfinalcopy 

\title{Extracting Implicitly Asserted Propositions in Argumentation}

\author{Yohan Jo$^1$ ~ Jacky Visser$^2$ ~ Chris Reed$^2$ ~ Eduard Hovy$^1$ \\
  $^1$Language Technologies Institute, Carnegie Mellon University, USA \\
  $^2$Centre for Argument Technology, University of Dundee, UK \\
  \texttt{yohanj@cs.cmu.edu},~ \texttt{j.visser@dundee.ac.uk}, \\ \texttt{c.a.reed@dundee.ac.uk},~ \texttt{hovy@cmu.edu} \\}

\date{}

\backgroundsetup{
    angle=0,
    scale=1,
    color=blue,
    position=current page.south,
    vshift=1.5cm,
    contents={\scriptsize \ifnum\value{page}=1 \shortstack[l]{@inproceedings\{Jo:2020implicit, \\
    \hspace{2cm}author = \{Jo, Yohan and Visser, Jacky and Reed, Chris and Hovy, Eduard\}, \\
    \hspace{2cm}title = \{Extracting Implicitly Asserted Propositions in Argumentation\}, \\
    \hspace{2cm}booktitle = \{Proceedings of the 2020 Conference on Empirical Methods in Natural Language Processing\}, year = \{2020\}\}} \fi}
}

\begin{document}
\maketitle
\begin{abstract}
    Argumentation accommodates various rhetorical devices, such as questions, reported speech, and imperatives. These rhetorical tools usually assert argumentatively relevant propositions rather implicitly, so understanding their true meaning is key to understanding certain arguments properly. However, most argument mining systems and computational linguistics research have paid little attention to implicitly asserted propositions in argumentation. In this paper, we examine a wide range of computational methods for extracting propositions that are implicitly asserted in questions, reported speech, and imperatives in argumentation. By evaluating the models on a corpus of 2016 U.S. presidential debates and online commentary, we demonstrate the effectiveness and limitations of the computational models. Our study may inform future research on argument mining and the semantics of these rhetorical devices in argumentation.\footnote{Our data and source code are available at \url{github.com/yohanjo/emnlp20_prop_extr}. All details for reproducibility are in Appendix \ref{sec:reproducibility}.}
\end{abstract}

\section{Introduction}
\label{intro}

Argument mining is a growing research field in computational linguistics. One of its main goals is to automatically identify pro- and counter-arguments underlying argumentative discourse. The foundational step for argument mining is to extract the elementary units of arguments in the discourse, after which the support or attack relations between these units are identified. According to argumentation theory, the elementary units in argumentation are \textit{asserted propositions}~\cite{vanEemeren:1984dp}. However, the dominant approach to extracting elementary units from text---often called \textit{argumentative discourse unit segmentation}~\cite{Ajjour:2017vm,Persing:2016cz,Jo:2019wu}---is rather simplistic and may even seem inconsistent with the theory. This approach segments text into smaller pieces (e.g., clauses) and treats each segment as an elementary unit of arguments. But these segments include locutions that are seemingly not assertives, such as questions and imperatives used as rhetorical devices. In fact, questions, imperatives, and reported speech in argumentation often assert propositions implicitly. Therefore, in order to understand certain argumentation and identify pro-/counter-arguments properly, locutions in argumentation should not be taken literally in their surface forms; instead, we need to go further and understand what propositions are implicitly asserted and argumentatively relevant in those locutions. Our work provides some computational solutions to this problem, namely, extracting implicitly asserted propositions in argumentation.

The following example dialogue illustrates how questions, reported speech, and imperatives assert propositions implicitly in argumentation. 
\begin{align}
    A: ~& \text{\lex{All human should be vegan.}} \label{ex:conclusion}\\
    & \text{\lex{Look at how unethical the meat}} \label{ex:imperative} \\[-0.4em]
    & \text{\lex{production industry is.}} \nonumber \\
    & \text{\lex{Environmental scientists proved that}} \label{ex:rspch} \\[-0.4em]
    & \text{\lex{vegan diets reduce meat production by 73\%.}} \nonumber\\
    B: ~& \text{\lex{Well, don't vegan diets lack essential}} \label{ex:question} \\[-0.4em]
    & \text{\lex{nutrients, though?}} \nonumber
\end{align}
In this dialogue, speaker \textit{A} is supporting conclusion \ref{ex:conclusion} using sentences \ref{ex:imperative} and \ref{ex:rspch}, whereas speaker \textit{B} is attacking the conclusion using sentence \ref{ex:question}. Sentence \ref{ex:imperative} is an imperative, but in this argumentation, it is \textit{asserting} that the meat production industry \textit{is} unethical. In sentence \ref{ex:rspch}, the primary proposition asserted in support of the conclusion is the content of this reported speech---``vegan diets reduce meat production by 73\%''; the ``environmental scientists'' is presented as the source of this content in order to strengthen the main proposition in this sentence. Lastly, sentence \ref{ex:question} is in question form, but it is in fact \textit{asserting} that vegan diets \textit{lack} essential nutrients. These examples suggest that properly understanding arguments requires comprehension of what is meant by questions, reported speech, and imperatives, that is, what they assert implicitly.

In this paper, we test various computational methods to extract propositions that are implicitly asserted in questions, reported speech, and imperatives. Across the tasks, we explore a wide range of computational methods. For questions, we develop neural and rule-based methods for transforming questions into asserted propositions. For reported speech, we present feature-based and neural models to identify speech content (the primary proposition asserted) and speech source. Lastly, for imperatives, we test a simple transformation rule manually and analyze the patterns of how they assert propositions. By evaluating our models on a corpus of the 2016 U.S. presidential debates and online commentary, we demonstrate their effectiveness and limitations. 

Our contributions are as follows:
\begin{itemize}
    \item Our work is a first computational study of extracting propositions asserted in questions, reported speech, and imperatives in argumentation. We demonstrate the effectiveness and limitations of various computational models. This problem is fundamental in argument mining, albeit understudied.
    \item We find the evidence of strong syntactic regularities in how propositions are asserted in question form. 
    \item We show the robust performance of a state-of-the-art language model for identifying speech content and source in reported speech.
    \item Our case study of how imperatives implicitly assert propositions is novel in computational linguistics and argumentation theory. This study may inform future research on the semantics of imperatives in argumentation.
\end{itemize}

\section{Background\label{sec:background}}
Argumentation is an illocutionary act of supporting or attacking an expressed opinion by \textit{asserting} propositions, according to Pragma-Dialectics~\cite{vanEemeren:1984dp}. This definition might seem counterintuitive, as argumentation often accommodates locutions that are not assertives, such as questions and imperatives. We will draw upon theory and discuss how propositions are asserted implicitly in questions, reported speech, and imperatives in argumentation. But for the sake of the readability of the paper, we will defer this discussion to the respective sections of questions (\S{\ref{sec:question}}), reported speech (\S{\ref{sec:rspch}}), and imperatives (\S{\ref{sec:imperative}}). 

On the other hand, one of the main goals of argument mining is to identify pro- and counter-relations between asserted propositions. In most argument mining systems, asserted propositions are approximated and substituted by argumentative discourse units (ADUs). An ADU is the minimal locution that performs an argumentative function. Given an utterance, ADUs may be identified based on syntactic rules, such as phrases~\cite{Stede:2016seg}, clauses~\cite{Peldszus:2015ku}, or a series of clauses~\cite{AlKhatib:2016news}, or by machine learning models, such as neural networks~\cite{Ajjour:2017vm} or retrieval~\cite{Persing:2016cz}. None of these methods go further to understand what propositions are asserted in each ADU.

More recently, a computational framework has been proposed to extract asserted propositions from ADUs~\cite{Jo:2019wu}. This cascade model proposes how to detect reported speech, questions, and imperatives, reconstruct any missing subjects, and make final revisions for grammar correction. While this model was built upon the same goal of extracting asserted propositions from locutions, it does not present computational models to extract implicit propositions in questions, reported speech, and imperatives. Hence, our work fills this gap in the cascade model.

\section{Domain}
The domain we focus on is 2016 U.S. presidential debates and online commentary on Reddit~\cite{Visser:2019un}. This corpus includes the first Republican candidates debate for the primaries, the first Democratic candidates debate for the primaries, and the first general election debate. The corpus also includes Reddit discussions on these debates. 

Each utterance has been segmented into ADUs, and each ADU has been further annotated with an asserted proposition. The inter-annotator agreement is Cohen's $\kappa$ of 0.61 (substantial agreement). These debates are ideal for our analysis, since they accommodate questions, reported speech, and imperatives from various speakers and in both formal and informal debate settings.

Our work uses the data pre-processed by \newcite{Jo:2019wu}. This dataset has resolved anaphors in ADUs and paired ADUs with asserted propositions in a readily-available format\footnote{\url{https://github.com/yohanjo/amw19}}. While most of our work is based on this dataset, individual tasks need additional processing or additional data. They will be described in the respective section.

\section{Questions\label{sec:question}}
In this section, we extract implicit propositions from questions in argumentation. The task is formulated as transforming a question into its asserted proposition. 

\subsection{Theoretical Background}
Questions in argumentation may be categorized into rhetorical questions and pure questions. Rhetorical questions are not intended to require an answer; instead, they often make an implicit assertive (as in sentence \ref{ex:question}). \newcite{Zhang:2017wv} identified finer-grained types of rhetorical questions, such as sharing concerns, agreeing, and conceding. Our work is not aiming to classify these types, but instead focuses on extracting implicit assertives in rhetorical questions.

Pure questions, on the other hand, are intended to seek information. According to the speech act theory, non-binary questions have incomplete propositions~\cite{Searle:1969cb}. For instance, the question ``\lex{How many people were arrested?}'' has the proposition ``\lex{X people were arrested}'', with the questioned part underspecified and denoted by \lex{X}. Although the proposition is semantically underspecified, subsequent arguments may build on this, making this proposition an important argumentative component. Hence, our work covers extracting semantically underspecified propositions from pure questions as well. (See \newcite{Bhattasali15:question} for computational methods to distinguish between rhetorical questions and pure questions.)

\subsection{Models}
We explore two neural seq2seq models and one rule-based model. For all these models, both input and output are a sequence of words.

\subsubsection{Neural Models}
\begin{figure*}[t]
    \centering
    \includegraphics[width=\linewidth]{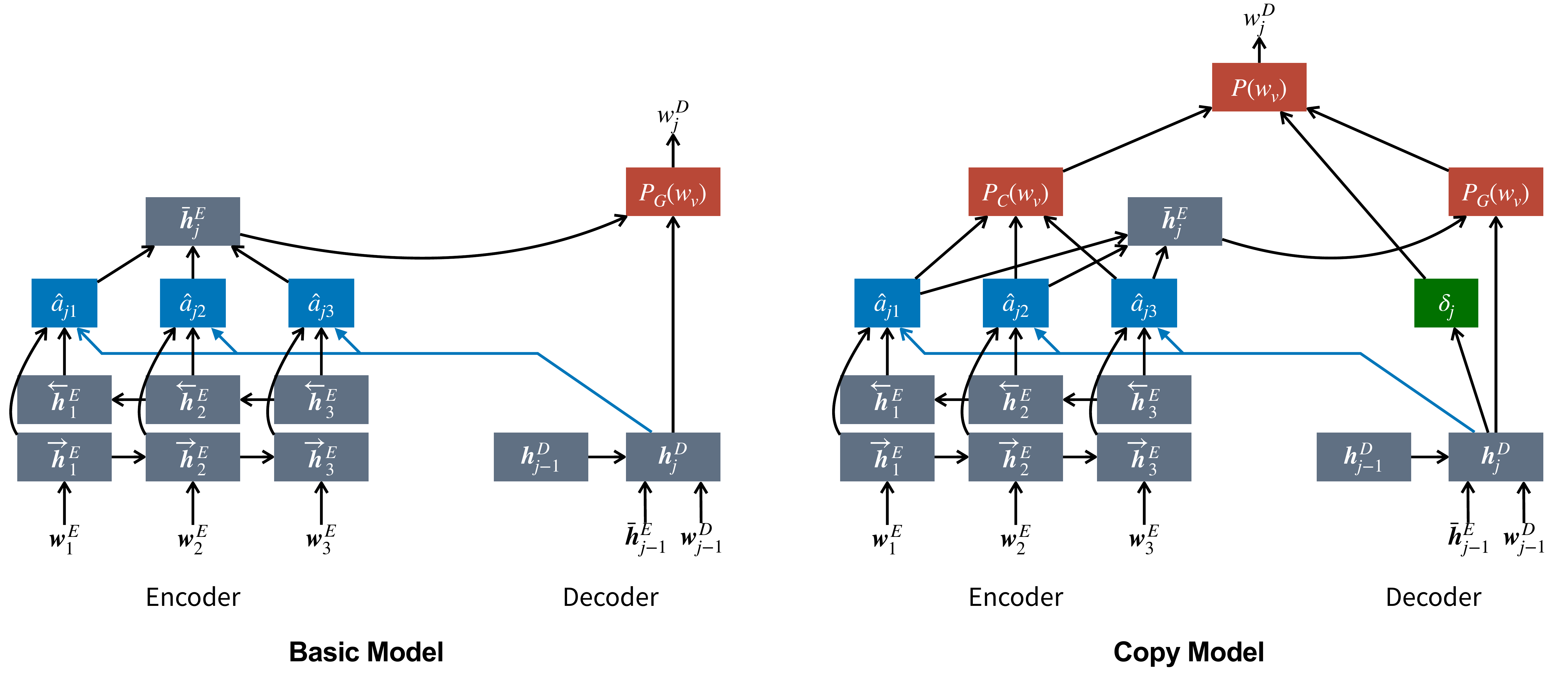}
    \caption{Basic model and copy model for question transformation. The snapshots for the $j$th output word.}
    \label{fig:quest_models}
\end{figure*}

\newcommand{\rhat}[1]{\overrightarrow{#1}}
\newcommand{\lhat}[1]{\overleftarrow{#1}}
\newcommand{\hv}{\boldsymbol{h}}
\newcommand{\wv}{\boldsymbol{w}}
\setlength\abovedisplayskip{1pt}
\setlength\belowdisplayskip{1pt}

We test two RNN-based seq2seq models. First, the \textbf{basic} model encodes a question using BiLSTM and decodes a proposition using LSTM and the standard attention mechanism~\cite{Luong:2015wx}. Figure \ref{fig:quest_models} illustrates the snapshot of the model for the $j$th output word.

Formally, the input is a sequence of words $w^E_1, \cdots, w^E_N$, and the embedding of $w^E_i$ is denoted by $\wv^E_i$. BiLSTM encodes each word $\wv^E_i$ and outputs forward/backward hidden states $\rhat{\hv}^E_i$ and $\lhat{\hv}^E_i$:
\begin{gather*}
    \rhat{\hv}^E_i, \lhat{\hv}^E_i = \textrm{BiLSTM}(\wv^E_i, \rhat{\hv}^E_{i-1}, \lhat{\hv}^E_{i+1}), \\
    \rhat{\hv}^E_0 = \lhat{\hv}^E_{N+1} = \boldsymbol{0}.
\end{gather*}
For the $j$th word to be generated, an LSTM decoder encodes the concatenation of the previously generated word $\wv^D_{j-1}$ and context vector $\bar{\hv}^E_{j-1}$ (explained below), and the previous hidden state:
\begin{gather*}
    \hv^D_j = \textrm{LSTM}([\wv^D_{j-1}; \bar{\hv}^E_{j-1}], \hv^D_{j-1}), \\
    \hv^D_0 = [\lhat{\hv}^E_1; \rhat{\hv}^E_N].
\end{gather*}
Next, the decoder attends to the encoder's hidden states using an attention mechanism. The attention weight of the $i$th hidden state is the dot product of the hidden states from the encoder and the decoder:
\begin{gather*}
    a_{ji} = \hv^{D}_j \cdot [\lhat{\hv}^E_i; \rhat{\hv}^E_i], 
    ~~\hat{a}_{ji} = \frac{\exp(a_{ji})}{\sum_{i'} \exp(a_{ji'})}, \\
    \bar{\hv}^E_j = \sum_i \hat{a}_{ji} [\rhat{\hv}^E_i; \lhat{\hv}^E_i].
\end{gather*}
The probability of the $v$th word in the vocabulary being generated is calculated as in the standard attention decoder mechanism:
\begin{gather*}
    P_{G}(w_v) = \textrm{softmax}(W_G [\hv^D_j; \bar{\hv}^E_j] + \boldsymbol{b}_G)_v,
\end{gather*}
where $W_G$ and $\boldsymbol{b}_G$ are trainable weight matrix and bias vector.

The basic seq2seq model requires a lot of training data, whereas according to our observation, question transformation is often formulaic, consisting largely of word reordering. Hence, our \textbf{copy} model uses a copying mechanism to learn to re-use input words. A prior model~\cite{Gu:2016jb} does not perform well in our task, so we modified it as follows (Figure \ref{fig:quest_models}).

Our copy model is based on the basic model and has the same process for the generating part. When an output word is copied from the input text, instead of being generated, the probability of the $i$th input word being copied is proportional to the attention weight of the $i$th hidden state. That is, the probability of the $v$th word in the vocabulary being copied is:
\begin{gather*}
    P_{C}(w_v) = \sum_{i=1}^{N} \hat{a}_{ji} I(w^E_i = w_v).
\end{gather*}

The final probability of $w_v$ being output is a weighted sum of $P_C(w_v)$ and $P_G(w_v)$, where the weight $\delta$ is calculated as
\begin{gather*}
    \delta_j = \sigma(W_\delta \hv^D_j + \boldsymbol{b}_\delta), \\
    P(w_v) = \delta P_{C}(w_v) + (1 - \delta)P_{G}(w_v),
\end{gather*}
where $W_\delta$ and $\boldsymbol{b}_\delta$ are trainable weight matrix and bias vector. The main difference of our model from existing ones is that we compute the mixture weight $\delta_j$ for $P_C$ and $P_G$ using a separate neural network. In contrast, existing models do not explicitly compute this weight~\cite{Gu:2016jb} or do not use attentional hidden states~\cite{Allamanis:2016tk}.

We try the following hyperparameter values:
\begin{itemize}
    \item Encoder/decoder hidden dim: 96, 128, 160, 192 (basic model) / 128, 192 (copy model)
    \item Beam size: 4
    \item Optimizer: Adam
    \item Learning rate: 0.001
    \item Gradient clipping: 1
    \item Word embedding: GloVe 840B
\end{itemize}

\subsubsection{Rule-Based Model}
As question transformation is often formulaic, a rule-based method may be effective for small data. For each question, the most relevant parts for transformation are the first word (wh-adverb or auxiliary verb), subject, auxiliary verb, negation, and main verb (i.e., \lex{be}+adjective, \lex{be}+gerund, or else). For instance, the question ``\lex{\underline{Why} \underline{would} \underline{you} \underline{not} \underline{pay} the tax?}'' might be rearranged to ``\lex{\underline{You} \underline{would} \underline{pay} the tax}'', where \lex{why} and \lex{not} are removed. We compile rules that match combinations of these components, starting with a rule that has a high coverage and breaking it down to more specific ones if the rule makes many errors. An example rule is ``\lex{Why} [MODAL] [SUBJECT] \lex{not}'' $\rightarrow$ ``[SUBJECT] [MODAL]'', which applies to the above example. As a result, we compiled total 94 rules for 21 first words (4.5 rules per first word on average) based on the US2016 dataset (see Table \ref{tab:quest_trans_rules} in Appendix \ref{sec:quest_trans_rules} for a summary of these rules).

\subsection{Data}
\paragraph{US2016:} Our main data is \newcite{Jo:2019wu}'s dataset of the 2016 U.S. presidential debates and commentary. 
We filtered 565 pairs of an ADU and its asserted proposition that are annotated with the following question types:
\begin{itemize}[nolistsep]
    \item \textbf{Pure:} e.g., \texttotext{``Who is Chafee?''}{``Chafee is xxx''}; \texttotext{``Do lives matter?''}{``Lives do / do not matter''} (Semantically underspecified parts are denoted by \lex{xxx} and the slash \lex{/}.)
    \item \textbf{Assertive:} e.g., \texttotext{``What does that say about your ability to handle challenging crises as president?''}{``Clinton does not have the ability to handle challenging crises as president''}
    \item \textbf{Challenge:} e.g., \texttotext{``What has he not answered?''}{``He has answered questions''}
    \item \textbf{Directive:} e.g., \texttotext{``Any specific examples?''}{``Provide any specific examples''}
\end{itemize}
Note that only pure questions are semantically underspecified (indicated by \lex{xxx} and \lex{/}); the other types contain concrete propositions to be asserted.
Our models are trained on all question types.

\paragraph{MoralMaze:} This dataset consists of 8 episodes of the BBC Moral Maze Radio 4 program from the 2012 summer season\footnote{\url{http://corpora.aifdb.org/mm2012}}~\cite{Lawrence15mm2012}. The episodes deal with various issues, such as the banking system, welfare state, and British empire. In each episode, the BBC Radio presenter moderates argumentation among four regular panelists and three guest participants. This dataset has been annotated in the same way as US2016, and we filtered 314 pairs of a question and its asserted proposition. This dataset is not used for training or compiling rules; instead, it is only used as a test set to examine the domain-generality of the models.

\subsection{Experiment Settings}
For the neural models, we conduct two sets of experiments. First, we train and test the models on US2016 using 5-fold cross validation. Second, to examine domain generality, we train the models on the entire US2016 dataset and test on MoralMaze. 

For the rule-based model, we compile the rules based on US2016 and test them on US2016 (previously seen) and MoralMaze (unseen). 

The accuracy of the models is measured in terms of the BLEU score, where the references are asserted propositions annotated in the dataset.

\subsection{Result}
\begin{table}[t]
    \small
    \centering
    \begin{tabularx}{\linewidth}{XS[table-format=3.1]cS[table-format=3.1]c} \toprule
         & \multicolumn{2}{c}{US2016} & \multicolumn{2}{c}{MoralMaze} \\
        \cmidrule(r){2-3} \cmidrule{4-5}
         & {BLEU} & \%M &  {BLEU} & \%M \\ 
        \cmidrule(r){1-3} \cmidrule{4-5}
        Original Questions & 47.5 & -- & 50.7 & -- \\
        Basic Model & 5.3 & -- & 6.5 & -- \\
        Copy Model & 41.5 & -- & 44.1 &  -- \\
        Rules & 54.5 & \hide{359/565} 64\% & 51.9 & \hide{149/314} 48\% \\
        Rules (well-formed) & 56.7 & \hide{262/309} 85\% & 54.5 & \hide{103/150} 69\% \\
        \bottomrule
    \end{tabularx}
    \caption{Accuracy of extracting implicitly asserted propositions from questions. ``\%M'' is the percentage of questions matched with any hand-crafted rules.}
    \label{tab:quest_trans_perf}
\end{table}

As shown in Table \ref{tab:quest_trans_perf}, the basic seq2seq model (row 2) performs poorly, because of the small size of the training data. On the other hand, the copy model (row 3) significantly improves the BLEU scores by 36.2--37.6 points, by learning to re-use words in input texts\footnote{Our model also outperforms a prior copy model~\cite{Gu:2016jb} by more than 20 BLEU scores.}. However, it still suffers the small data size, and its outputs are worse than the original questions without any transformation (row 1). 

In contrast, the hand-crafted rules (rows 4--5) significantly improve performance and outperform the original questions. The effectiveness of the rule-based method on MoralMaze, which was not used for compiling the rules, indicates that these rules generalize across argumentative dialogue\footnote{Yet, we do not believe these rules would be effective beyond argumentation if the distribution of rhetorical questions and pure questions is significantly different from argumentative dialogue.}. The effectiveness of the rule-based method also suggests that there exist a high degree of syntactic regularities in how propositions are asserted implicitly in question form, and the hand-crafted rules provide interpretable insights into these regularities (see Table \ref{tab:quest_trans_rules} in Appendix \ref{sec:quest_trans_rules} for the rules).

Taking a closer look at the rule-based method, we find that many questions are subordinated or ill-formed, and thus the rules match only 64\% of questions for US2016 and 48\% of questions for MoralMaze. When we focus only on well-formed questions (that begin with a wh-adverb or auxiliary verb), the rules match 85\% and 69\% of questions for the respective dataset, and the BLEU scores improve by 2.2--2.6 points (row 4 vs. row 5). When analyzed by the first word of a question, questions beginning with \lex{have}, \lex{do}, and modal verbs achieve the highest BLEU scores. Why-questions achieve the lowest, probably due to many variants possible; for example, ``\lex{why isn't} [SUBJECT] [ADJECTIVE]\lex{?}'' is most likely to be transformed to ``[SUBJECT] \lex{is} [ADJECTIVE]'', whereas ``\lex{why isn't} [SUBJECT] [VERB]\lex{?}'' is to ``[SUBJECT] \lex{should be} [VERB]''.

One limitation of the rule-based method, however, is that it cannot distinguish between questions that have the same syntactic structure but assert opposite propositions. For example, ``\lex{Would you ...?}'' can mean both ``\lex{You would ...}'' and ``\lex{You would not ...}'' depending on the context. In order to separate these cases properly, we may need to take into account more nuanced features and context, and machine learning with large data would be the most promising direction eventually.

\section{Reported Speech\label{sec:rspch}}
In this section, we extract speech content (i.e., propositions that are often asserted as the primary contribution to the argumentation) and speech source in reported speech. This task is formulated as sequence tagging: words that constitute speech content or source are tagged with B followed by I, and all other words are tagged with O. 

\subsection{Theoretical Background}
Reported speech consists of \textit{speech content} that is borrowed from a \textit{speech source} external to the speaker. Speech content can be a direct quote of the original utterance or an indirect, possibly paraphrased utterance. Reported speech is a common rhetorical device in argumentation and performs various functions, including:
\begin{itemize}
    \item Appeals to authority by referencing experts or rules~\cite{Walton:2008schem} (e.g., ``\lex{Environmental scientists proved that vegan diets reduce meat production by 73\%.}'')
    \item Sets a stage for dis/agreeing with the position~\cite{Janier:2017jj} (e.g., ``\lex{You say that you want attention, but, at the same time, you don’t want me to bring attention to you.}'')
    \item Commits straw man fallacies by distorting the original representation or selecting part of the original utterance~\cite{Talisse:2006ck}
\end{itemize}
While reported speech as a whole is an assertion, its primary contribution to the argumentation usually comes from the speech content (e.g., ``vegan diets reduce meat production by 73\%''), and the speech source (e.g., ``environmental scientists'') is used to support the speech content.

Due to the important roles of speech content and source, computational models have been proposed to identify them, based on rules~\cite{Krestel08:rspch}, conditional random fields~\cite{Pareti03:quotation}, and a semi-Markov model~\cite{Scheible16:quotation}. Our work is different from these studies in two ways. First, they are based on news articles, whereas our work is on argumentative dialogue. Second, they use rules or features that reflect typical words and structures used in reported speech, whereas our work explores a neural method that does not require feature engineering. We aim to show how well a state-of-the-art neural technique performs on extraction of speech content and source. A slightly different but related strain of work is to identify authority claims in Wikipedia discussions~\cite{Bender:2011ud}, but this work does not identify speech content and source.

\subsection{Models}
We explore three models: a conditional random field (CRF) with hand-crafted features, the BERT token classifier with a pretrained language model, and a semi-Markov model as the baseline. For all models, the input is a sequence of words and the output is a BIO tag for each word.
We conduct separate experiments for content and source, because we do not assume that they are mutually exclusive (although they are in most cases).

\subsubsection{Conditional Random Field (CRF)} 
Our CRF uses the following features:
\begin{itemize}
    \item Current word.
    \item Named entity type of the word.
    \item POS tag of the word.
    \item Unigram and bigram preceding the word.
    \item Unigram and bigram following the word.
    \item Indicator of if the word is a subject (``nsubj*'' on the dependency parse tree).
    \item Indicator of if the current word is the beginning/end of a clause (``S'' on the parse tree).
\end{itemize}
The features were extracted using Stanford CoreNLP 0.9.2~\cite{Manning:2014corenlp}.

For model parameters, we explore two optimization functions: (i) L-BFGS with the combinations of L1/L2 regularization coefficients $\{0, .05, .1, .2\}$; (ii) Passive Aggressive with aggressiveness parameter values $\{.5, 1, 2, 4\}$. The model was implemented using sklearn\_crfsuite 0.3.6. 

\subsubsection{BERT} 
The second model is the BERT token classifier~\cite{Devlin:2018bert}, which classifies the tag of each word. BERT has shown significant performance boosts in many NLP tasks and does not require hand-crafted features. 
We use the pretrained, uncased base model with the implementation provided by Hugging Face~\cite{huggingface2019}. The model is fine-tuned during training.

\subsubsection{Baseline\label{sec:rspch_baseline}}
The baseline is the state-of-the-art semi-Markov model for speech content identification~\cite{Scheible16:quotation}. This model first identifies cue words (e.g., reporting verbs) and iteratively identifies the boundaries of speech content using a set of hand-crafted features. This model does not identify speech sources and thus is compared with other models only for content identification. 

For a methodological note, the original source code was hard-coded to work for the PARC3.0 dataset, and we could not replicate the model to train on other data. Therefore, all accuracies of this model in the next section result from training it on the training set of the PARC3.0 dataset (Section \ref{sec:rspch_data}). We will show its performance on both PARC3.0 and US2016.

\subsection{Data\label{sec:rspch_data}}
\paragraph{PARC3.0:} The first dataset is 18,201 instances of reported speech in news data~\cite{Pareti:2016parc}. The original dataset was built upon the Wall Street Journal articles in the Penn Discourse TreeBank (PDTB)~\cite{Prasad:2008pdtb}, where each instance of reported speech has been annotated with the content, source, and cue word (e.g., reporting verbs). The reliability of the annotations were measured by the overlap of annotated text spans between annotators. The overlap for speech content is 94\% and that for speech source is 91\%, suggesting the high reliability of the annotations.

This dataset consists of 24 sections corresponding to the PDTB sections. The original paper suggests using sections 00-22 for training (16,370 instances), section 23 for testing (667 instances), and section 24 for validation (1,164 instances).

\paragraph{US2016:} The second dataset is the instances of reported speech in the corpus of the 2016 U.S. presidential debates and commentary, prepared by \newcite{Jo:2020lrec}\footnote{\url{https://github.com/yohanjo/lrec20}}. This dataset includes 242 instances of reported speech annotated with speech content and source. The reliability of the annotations was measured by the number non-overlapping words between annotators. The average number of words that are outside of the overlapping text span was 0.2 for speech content and 0.5 for speech sources, suggesting the high reliability of the annotations.

\subsubsection{Experiment Settings}
The CRF and BERT models are trained and tested on both PARC3.0 and US2016, separately. For PARC3.0, we use the split of train, validation, and test as suggested by the original paper.  For US2016, we use 5-fold cross validation; for each iteration, three folds are used for training, one for testing, and the other for choosing the optimal hyperparameters (CRF) or the optimal number of epochs (BERT). 

The baseline model is trained and tested on PARC3.0 using the same training, validation, and test split. US2016 is used only for testing after it is trained on the training set of PARC3.0 (as mentioned in \ref{sec:rspch_baseline}).

We use various evaluation metrics. For speech content, the \textbf{F1-score} is calculated based on the true and predicted BIO tags of individual words, as well as the \textbf{BLEU} score of the predicted text span against the true text span.
For speech sources, the F1-score is calculated based on the match between the true source's text and the predicted text. Two texts are considered matched if they are identical (\textbf{Strict}) or if their words overlap (\textbf{Relaxed}). We do not measure the F1-score based on BIO tags for speech sources, because the source may be mentioned multiple times in reported speech and we do not want to penalize the model when the mention identified by the model is the true source but different from the annotated mention.

\subsection{Result}

\begin{table}[t]
    \small
    \centering
    \begin{subtable}[t]{\linewidth}
        \begin{tabularx}{\linewidth}{Xcccc}\toprule
             & \multicolumn{2}{c}{PARC3.0} & \multicolumn{2}{c}{US2016} \\
             \cmidrule(r){2-3} \cmidrule(l){4-5}
             & F1 & BLEU & F1 & BLEU  \\ \cmidrule(r){1-3} \cmidrule(l){4-5}
            Scheible (All) & 64.4 & 57.1 & \textit{37.9} & \textit{23.4}  \\
            Scheible (Matched) & 75.8 & 72.7 & \textit{79.3} & \textit{76.5} \\
            \midrule
            CRF & 71.3 & 66.3 & 72.5 & 68.7 \\
            BERT & 82.6 & 82.0 & 87.1 & 89.3 \\ 
            \bottomrule
        \end{tabularx}
        \caption{Accuracy of identifying speech content. The accuracies of Scheible for US2016 (italic) result from training it on the training data of PARC3.0.}
        \label{tab:spch_ident}
    \end{subtable}
    \begin{subtable}[t]{\linewidth}
        \begin{tabularx}{\linewidth}{Xcccc}\toprule
             & \multicolumn{2}{c}{PARC3.0} & \multicolumn{2}{c}{US2016} \\
             \cmidrule(r){2-3} \cmidrule(l){4-5}
             & Strict F1 & Relaxed F1 & Strict F1 & Relaxed F1  \\ \cmidrule(r){1-3} \cmidrule(l){4-5}
            CRF & 52.4 & 59.8 & 62.4 & 71.6  \\
            BERT & 71.0 & 78.6 & 70.3 & 84.8  \\ 
            \bottomrule
        \end{tabularx}
        \caption{Accuracy of identifying speech source.}
        \label{tab:spkr_ident}
    \end{subtable}
    \caption{Accuracy of identifying speech content and source.}
    \label{tab:spch_spkr_ident}
\end{table}

\paragraph{Content Identification:} 
The accuracies of all models are summarized in Table \ref{tab:spch_ident}. The baseline model (Scheible) has two rows: row 1 is its accuracy on all test instances, and row 2 is on test instances where the model was able to identify cue words. We find that the BERT model (row 4) outperforms the feature-based CRF and the baseline model for both corpora, achieving a macro F1-score of 82.6\% at tag levels and a BLEU score of 82.0\% for PARC3.0 and an F1-score of 87.1\% and a BLEU score of 89.3\% for US2016. These scores show the high reliability of the BERT model for extracting main propositions asserted in reported speech. In addition, the high accuracy on US2016 despite its small size suggests that the pretrained language model effectively encodes important semantic information, such as reporting verbs and dependencies among subject, verb, and object.

The baseline model, which was trained on PARC3.0, performs poorly on US2016 (row 1). The main obstacle is that it fails to detect cue words (e.g., reporting verbs) in 168 out of 242 instances (69\%). This shows one weakness of the baseline model: since this model works at two steps---detect cue words and find content boundaries---identifying speech content is strongly subject to cue word detection. When the baseline is evaluated only on the instances where a cue word was detected, its accuracy boosts significantly (row 2), outperforming the CRF but still worse than BERT. 

A qualitative analysis of the BERT model reveals that most instances are tagged accurately, and errors are concentrated on a few instances. One of the main issues is whether a reporting verb should be included or not as speech content. In the annotation process for US2016, a reporting verb was included as speech content only if the verb has meaning other than merely ``to report'' (e.g., \lex{\textbf{blamed} his idea}, \lex{\textbf{declared} their candidacy}). As a result, the model often has difficulty judging a reporting verb to be part of the speech content or not.

In some cases, the exact boundary of speech content is ambiguous. For instance, in the sentence
\begin{quote}
    ``\lex{Bush has promised \ul{\textbf{four percent economic growth and 19 million new jobs} if Bush is fortunate enough to serve two terms as president}.}''
\end{quote}
the annotated speech content is in bold, while the model included the if-clause as the content (underlined). However, it may seem more appropriate to include the if-clause as part of the promise.

\paragraph{Source Identification:} 
The accuracies of all models are summarized in Table \ref{tab:spkr_ident}. The BERT model (row 2) again significantly outperforms the CRF (row 1), achieving F1-scores of 75.7\% for strict evaluation (exact match) and 85.1\% for relaxed evaluation (overlap allowed). 
It is usually when a source is a long noun phrase that a predicted source and the true source overlap without exact match (e.g., \lex{President Obama} vs. \lex{Obama}).

Our qualitative analysis of the BERT model reveals two common error cases. First, the model tends to capture subjects and person names as a speech source, which is not correct in some cases:
\begin{quote}
``\lex{We have been told through investigative reporting that he owes about \$650 million to Wall Street and foreign banks}''    
\end{quote}
where the model identifies \lex{we} as the speech source, while the true source is the \lex{investigative reporting}. The model also sometimes fails to detect any source candidate if reported speech has an uncommon structure, such as ``\lex{The \ul{record} shows that ...}'' and ``\lex{No one is arguing ... except for \ul{racists}}'', where the speech sources are underlined. These problems may be rectified with larger training data that include more diverse forms of reported speech.

\section{Imperatives\label{sec:imperative}}
In this section, we collect imperatives in argumentative dialogue and examine a simple method for extracting propositions asserted in them. 
We do not build automated models for transformation (as in questions), because US2016 had no clear guidelines on how to annotate asserted propositions in imperatives when the dataset was built.

\subsection{Theoretical Background}
Imperatives are common in argumentation as in ``\lex{Stop raising the sales tax}'' and ``\lex{Look how bad the system is}''. 
However, to our knowledge, there is little theoretical work on what propositional content is asserted by imperatives in argumentation. There have been theories about the semantics of imperatives in general context; for example, the \textit{you-should} theory suggests that an imperative of the form ``\lex{Do X}'' may imply ``\lex{X should be done}''~\cite{Hamblin.1987,Schwager.2005thesis}. 
While applicable in many general cases, this mechanism is not satisfactory in argumentation.
For instance, while this transformation preserves the literal meaning of both the first and second examples above, it does not capture the main proposition asserted in the second example. This example is unlikely arguing for ``looking'' per se; it rather asserts that the system is bad, which is the main content that contributes to the argumentation. No simple transformation rules apply here, and such irregularities call for more case studies. Our work aims to make an initial contribution in that direction.

\subsection{Model}
No automated model is used in this section, but instead, we examine the applicability of the \textit{you-should} theory in argumentation. Specifically, we analyze whether each imperative preserves the original intent when it is transformed to an assertive by adding ``\lex{should}'', along with appropriate changes in the verb form, (implicit) subject, and object. We additionally analyze the argumentative relevancy of the transformed verb, that is, whether the imperative is mainly asserting that it should happen.

\subsection{Data}
We use imperatives in US2016~\cite{Jo:2019wu}. We assume that a sentence is an imperative if its root is a verb in the bare infinitive form and has no explicit subject. Using Stanford CoreNLP, we chose locutions that are not questions and whose root is a verb with base form or second-person present case (VB/VBP), neither marked (e.g., \lex{to go}) nor modified by an auxiliary modal verb (e.g., \lex{would go}). We found total 191 imperatives, and the most common root verbs are listed in Table \ref{tab:imperatives}.

\subsection{Result}
\begin{table}[t]
    \small
    \centering
    \begin{tabularx}{\linewidth}{CCCC}\toprule
        Top 1-8 & Top 9-16 & Top 17-24 & Top 25-32  \\ \midrule
        let (39) & fuck (5) & say (3) & bring (2) \\
        look (7) & stop (5) & ask (2) & love (2) \\
        have (7) & do (4) & vote (2) & drink (2) \\
        wait (6) & check (3) & help (2) & pay (2) \\
        thank (6) & give (3) & keep (2) & are (2) \\
        please (6) & make (3) & find (2) & believe (2) \\
        go (5) & get (3) & think (2) & talk (2) \\
        take (5) & use (3) & forget (2) & screw (2) \\
        \bottomrule
    \end{tabularx}
    \caption{Root verbs and counts in imperatives.}
    \label{tab:imperatives}
\end{table}

We found that 74\% of the imperatives can be transformed to an assertion by adding \lex{should} while preserving their original meaning\footnote{Many of the other cases are attributed to subject drop (e.g., ``\lex{Thank you}'', ``\lex{Doesn't work}'') and CoreNLP errors (e.g., ``\lex{Please nothing on abortion}'', ``\lex{So do police jobs}'').}. And 80\% of the transformed assertions were found to be argumentatively relevant content. For example, the imperative ``\lex{Take away some of the pressure placed on it}'' can be transformed to (and at the same time asserts that) ``\lex{some of the pressure placed on it should be taken away}''. This result suggests that we can apply the \textit{you-should} theory to many imperatives and extract implicitly asserted propositions in consistent ways.

Some imperatives were found to be rather rhetorical, and the propositions they assert cannot be obtained simply by adding \lex{should}. Those imperatives commonly include such verbs as \lex{let}, \lex{fuck}, \lex{look}, \lex{wait}, and \lex{have}. The verb \lex{let} can assert different things. For instance, ``\lex{Let's talk about the real issues facing america}'' asserts that ``\lex{there are real issues facing america}'', while ``\lex{Let's solve this problem in an international way}'' asserts that ``\lex{we should solve this problem in an international way}''. The words \lex{fuck} and \lex{screw} are used to show strong disagreement and often assert that something should go away or be ignored. 

We cannot apply the same transformation rule to the same verb blindly, as a verb can be argumentatively relevant sometimes and only rhetorical at other times depending on the context. For instance, the verb \lex{take} in the above example is argumentatively relevant, but it can also be used only rhetorically as in ``\lex{Take clean energy (as an example)}''. 

Based on our analyses, we propose rough two-step guidelines for annotating propositions that are implicitly asserted in imperatives. First, we may group imperatives by their semantics based on theories, such as \textit{you-should} and \textit{you-will}~\cite{Schwager.2005thesis}. Second, for these imperatives, we may annotate whether the root verb is argumentatively relevant. For instance, if the \textit{you-should} theory is applicable to an imperative, we may annotate whether its verb is at the core of the main argumentative content that the speaker asserts should happen; the assertive form of this imperative is likely to be a statement that proposes a policy or action~\cite{Park:2018wy}. Argumentatively relevant imperatives may be annotated with asserted propositions using predefined transformation templates appropriate for their semantics. 
On the other hand, argumentatively irrelevant verbs may simply be rhetorical and need to be replaced properly. Annotation of these imperatives should handle many irregular cases, relying on the domain of the argumentation and the annotator's expertise.

\section{Conclusion}
Identifying implicitly asserted propositions in argumentation is key to understanding arguments properly. We presented and tested computational methods for extracting implicit propositions from questions and reported speech in argumentation. For transforming questions to propositions, hand-crafted rules were significantly more effective than neural models and provided insights into the regularities in how propositions are implicitly asserted in question form. Since rule-based methods do not take context into account, however, more annotated data would be needed for better question transformation based on machine learning. For reported speech, BERT-based models demonstrated high effectiveness in identifying speech content and source by utilizing the rich semantic information in the pretrained model.  Lastly, for imperatives, we demonstrated some regularities and irregularities in how propositions are asserted in imperatives. We find evidence that some verbs may need to be treated specially, while many other verbs could be treated in consistent ways.

\section*{Acknowledgments}
This research was supported by the Kwanjeong Educational Foundation and by UK EPSRC grant EP/N014871/1.

\bibliography{emnlp2020}
\bibliographystyle{acl_natbib}

\newpage

\appendix

\onecolumn

\section{Reproducibility Checklist\label{sec:reproducibility}}

Model settings for extracting implicit propositions from questions (Table \ref{tab:quest_trans_perf})
\begin{table}[!hp]
    \centering
    \small
    \begin{tabularx}{\linewidth}{p{3.7cm}CCCC} 
        \toprule
         & \multicolumn{2}{c}{Basic} & \multicolumn{2}{c}{Copy} \\ 
        \cmidrule(l){2-3} \cmidrule(l){4-5} \\
        Criterion & US2016 & MoralMaze & US2016 & MoralMaze \\
        \midrule
        Computing infrastructure & \multicolumn{4}{c}{\blap{Intel(R) Core(TM) i7-8700K CPU @ 3.70GHz / 31GiB System memory\\/ NVIDIA GP102 [TITAN Xp]}} \hide{check nairobi}  \\
        Number of parameters & 4,680,010 & 3,248,580 & 4,680,203 & 3,248,773 \\
        Validation performance & BLEU=10.7 \hide{check} & BLEU=11.6 \hide{check} & BLEU=47.1 & BLEU=49.7  \\
        \midrule
        Encoder/decoder hidden dim & \{96, 128, 160, 192\} & 192 & \{128, 192\} & 192 \\
        Other hyperparameters & \multicolumn{4}{c}{\blap{Beam size: 4\\ Optimizer: Adam\\ Learning rate: 0.001\\ Gradient clipping: 1\\ Word embedding: GloVe 840B}}  \\
        Optimal encoder/decoder hidden dim & 192 & 192 & 192 & 192 \\ 
        Number of hyperparameter search trials & 4 & (No hyperparameter search) & 2 & (No hyperparameter search)  \\
        Method of choosing hyperparameter values & \multicolumn{4}{c}{Grid search}  \\
        Criterion for selecting optimal hyperparameter values & \multicolumn{4}{c}{BLEU}  \\
        \bottomrule
    \end{tabularx}
    \caption{Reproducibility checklist for question transformation (Table~\ref{tab:quest_trans_perf}).}
    \label{tab:reproducibility_checklist}
\end{table}

\newpage

Model settings for extracting speech content from reported speech (Table \ref{tab:spch_ident})
\begin{table}[!hp]
    \centering
    \small
    \begin{tabularx}{\linewidth}{p{3.5cm}CCCC} 
        \toprule
         & \multicolumn{2}{c}{CRF} & \multicolumn{2}{c}{BERT} \\ 
        \cmidrule(l){2-3} \cmidrule(l){4-5} \\
        Criterion & PARC3.0 & US2016 & PARC3.0 & US2016 \\
        \midrule
        Computing infrastructure & \multicolumn{2}{c}{\blap{3.1 GHz Dual-Core Intel Core i7\\ / 16 GB 1867 MHz DDR3}} & \multicolumn{2}{c}{\blap{Intel(R) Core(TM) i7-8700K CPU\\ @ 3.70GHz\\ / 31GiB System memory\\/ NVIDIA GP102 [TITAN Xp]}} \\
        Average runtime & 17.6 mins  & 0.03 mins & 314.6 mins & 11.9 mins \\
        Number of parameters & 173,749 & 7,569 & \multicolumn{2}{c}{108M} \\
        Validation performance & F1=75.7, BLEU=72.2 & F1=75.6, BLEU=72.5 & F1=84.4, BLEU=83.8 & F1=88.1, BLEU=90.4 \\
        \midrule
        Bounds for hyperparameters & \multicolumn{2}{c}{\blap{(i) Optimization function: L-BFGS,\\ L1/L2 regularization\\ coefficients: $\{0, .05, .1, .2\}$\\ (ii) Optimization function:\\ Passive Aggressive,\\ Aggressive parameter values:\\ $\{.5, 1, 2, 4\}$}} & \multicolumn{2}{c}{\blap{Learning rate: 1e-5,\\ Adam $\epsilon$: 1e-8}} \\
        Optimal hyperparameter configuration & L-BFGS + L1=0.1 + L2=0.2 & L-BFGS + L1=0.05 + L2=0.1 & \multicolumn{2}{c}{Learning rate=1e-5 + Adam $\epsilon$=1e-8} \\
        Number of hyperparameter search trials & \multicolumn{2}{c}{20} & \multicolumn{2}{c}{(No hyperparameter search)} \\
        Method of choosing hyperparameter values & \multicolumn{2}{c}{Grid search} & \multicolumn{2}{c}{(No hyperparameter search)} \\
        Criterion for selecting optimal hyperparameter values & \multicolumn{2}{c}{F1} & \multicolumn{2}{c}{(No hyperparameter search)} \\
        \bottomrule
    \end{tabularx}
    \caption{Reproducibility checklist for extracting speech content from reported speech (Table~\ref{tab:spch_ident}).}
    \label{tab:reproducibility_checklist_rspch_content}
\end{table}

\newpage

Model settings for extracting speech source from reported speech (Table \ref{tab:spkr_ident})
\begin{table}[!hp]
    \centering
    \small
    \begin{tabularx}{\linewidth}{p{3.5cm}CCCC} 
        \toprule
         & \multicolumn{2}{c}{CRF} & \multicolumn{2}{c}{BERT} \\ 
        \cmidrule(l){2-3} \cmidrule(l){4-5} \\
        Criterion & PARC3.0 & US2016 & PARC3.0 & US2016 \\
        \midrule
        Computing infrastructure & \multicolumn{2}{c}{\blap{3.1 GHz Dual-Core Intel Core i7\\ / 16 GB 1867 MHz DDR3}} & \multicolumn{2}{c}{\blap{Intel(R) Core(TM) i7-8700K CPU\\ @ 3.70GHz\\ / 31GiB System memory\\/ NVIDIA GP102 [TITAN Xp]}} \\
        Average runtime & 12.6 mins & 0.02 mins & 314.7 mins & 15.7 mins  \\
        Number of parameters & 289,631 & 7,250 & \multicolumn{2}{c}{108M} \\
        Validation performance & Strict F1=61.7, Relaxed F1=67.8 & Strict F1=68.3, Relaxed F1=74.6 & Strict F1=75.0, Relaxed F1=80.7 & Strict F1=76.3, Relaxed F1=89.1 \\
        \midrule
        Bounds for hyperparameters & \multicolumn{2}{c}{\blap{(i) Optimization function: L-BFGS,\\ L1/L2 regularization\\ coefficients: $\{0, .05, .1, .2\}$\\ (ii) Optimization function:\\ Passive Aggressive,\\ Aggressive parameter values:\\ $\{.5, 1, 2, 4\}$}} & \multicolumn{2}{c}{\blap{Learning rate: 1e-5,\\ Adam $\epsilon$: 1e-8}} \\
        Optimal hyperparameter configuration & Passive Aggressive + Aggressive=1 & L-BFGS + L1=0 + L2=0.2 & \multicolumn{2}{c}{Learning rate=1e-5 + Adam $\epsilon$=1e-8} \\
        Number of hyperparameter search trials & \multicolumn{2}{c}{20} & \multicolumn{2}{c}{(No hyperparameter search)} \\
        Method of choosing hyperparameter values & \multicolumn{2}{c}{Grid search} & \multicolumn{2}{c}{(No hyperparameter search)} \\
        Criterion for selecting optimal hyperparameter values & \multicolumn{2}{c}{Strict F1} & \multicolumn{2}{c}{(No hyperparameter search)} \\
        \bottomrule
    \end{tabularx}
    \caption{Reproducibility checklist for extracting speech source from reported speech (Table~\ref{tab:spkr_ident}).}
    \label{tab:reproducibility_checklist_rspch_source}
\end{table}

\newpage

\section{Question Transformation Rules\label{sec:quest_trans_rules}}
\begin{table}[!hp]
    \small
    \centering
    \begin{tabularx}{.8\linewidth}{ll}
        \toprule
        From & To \\
        \midrule
        why [MD]$_1$ [SBJ]$_2$ [*]$_3$? &  [SBJ]$_2$ [MD]$_1$ not [*]$_3$. \\
        why [MD]$_1$ not [SBJ]$_2$ [*]$_3$? &  [SBJ]$_2$ [MD]$_1$ [*]$_3$. \\
        why do [SBJ]$_1$ [*]$_2$? &  [SBJ]$_1$ [*]$_2$. \\
        why [does$|$did]$_1$ [SBJ]$_2$ [*]$_3$? &  [SBJ]$_2$ [does$|$did]$_1$ [*]$_3$. \\
        why is [SBJ]$_1$ [*]$_2$? &  [SBJ]$_1$ is [*]$_2$ because xxx. \\
        why [are$|$were$|$was]$_1$ [SBJ]$_2$ [*]$_3$? &  [SBJ]$_2$ [are$|$were$|$was]$_1$ [*]$_3$. \\
        why [is$|$are$|$am]$_1$ not [SBJ]$_2$ [ADJ]$_3$? &  [SBJ]$_2$ [is$|$are$|$am]$_1$ [ADJ]$_3$. \\
        why [is$|$are$|$am]$_1$ not [SBJ]$_2$ [VP]$_3$? &  [SBJ]$_2$ should be [VP]$_3$. \\
        why not [VP]$_1$? &  should [VP]$_1$. \\
        \midrule
        where [do$|$did$|$does$|$MD]$_1$ [SBJ]$_2$ [*]$_3$? &  [SBJ]$_2$ [do$|$did$|$does$|$MD]$_1$ [*]$_3$ at xxx. \\
        when [did$|$has]$_1$ [SBJ]$_2$ [*]$_3$? &  [SBJ]$_2$ [did$|$has]$_1$ not [*]$_3$. \\
        \midrule
        how can [SBJ]$_1$ [*]$_2$? &  [SBJ]$_1$ cannot [*]$_2$. \\
        how [MD\textbackslash{}can]$_1$ [SBJ]$_2$ [*]$_3$? &  [SBJ]$_2$ [MD\textbackslash{}can]$_1$ [*]$_3$ by xxx. \\
        how [do$|$does]$_1$ [SBJ]$_2$ [*]$_3$? &  [SBJ]$_2$ [*]$_3$ by xxx. \\
        how [MD$|$do$|$does$|$did]$_1$ [SBJ]$_2$ not [*]$_3$? &  [SBJ]$_2$ should [*]$_3$. \\
        how are [SBJ]$_1$ going to [*]$_2$? &  [SBJ]$_1$ need to [*]$_2$. \\
        how are [SBJ]$_1$ supposed to [*]$_2$? &  [SBJ]$_1$ cannot [*]$_2$. \\
        how [am$|$are$|$is]$_1$ [SBJ]$_2$ not [*]$_3$? &  [SBJ]$_2$ should be [*]$_3$. \\
        how much [*]$_1$? &  xxx [*]$_1$. \\
        how [ADJ$|$ADV]$_1$ [VB$|$MD]$_2$ [SBJ]$_3$ [VP]$_4$? &  [SBJ]$_3$ [VB$|$MD]$_2$ [VP]$_4$. \\
        \midrule
        what [MD$|$did]$_1$ [SBJ]$_2$ [VB]$_3$ [*]$_4$? &  [SBJ]$_2$ [MD$|$did]$_1$ [VB]$_3$ xxx [*]$_4$. \\
        what [does$|$do]$_1$ [SBJ]$_2$ [VB]$_3$ [*]$_4$? &  [SBJ]$_2$ [VB]$_3$ xxx [*]$_4$. \\
        what am [SBJ]$_1$ [VB]$_2$ [*]$_3$? &  [SBJ]$_1$ am [VB]$_2$ xxx [*]$_3$. \\
        what [is$|$was$|$are]$_1$ [SBJ]$_2$? &  [SBJ]$_2$ [is$|$was$|$are]$_1$ xxx. \\
        what [VB\textbackslash{}did$|$does$|$do$|$am$|$was$|$is$|$are]$_1$ [*]$_2$? &  xxx [VB\textbackslash{}did$|$does$|$do$|$am$|$was$|$is$|$are]$_1$ [*]$_2$. \\
        \midrule
        which [*\textbackslash{}VB]$_1$ [*]$_2$? &  [*\textbackslash{}VB]$_1$ xxx. \\
        which [*\textbackslash{}VB]$_1$ [VB]$_2$ [SBJ]$_3$ [*]$_4$? &  [SBJ]$_3$ [VB]$_2$ [*]$_4$ [*\textbackslash{}VB]$_1$ xxx. \\
        \midrule
        who [VB]$_1$ [SBJ]$_2$ [VP]$_3$? &  [SBJ]$_2$ [VB]$_1$ [VP]$_3$ xxx. \\
        who is [SBJ]$_1$? &  [SBJ]$_1$ is xxx. \\
        who is [VP]$_1$? &  xxx is [VP]$_1$. \\
        who [*\textbackslash{}is]$_1$ [*]$_2$? &  xxx [*\textbackslash{}is]$_1$ [*]$_2$. \\
        \midrule
        have you not [*]$_1$? &  you have not [*]$_1$. \\
        {}[have$|$has]$_1$ [SBJ\textbackslash{}you]$_2$ [*]$_3$? &  [SBJ\textbackslash{}you]$_2$ [have$|$has]$_1$ [*]$_3$. \\
        is [SBJ]$_1$ [NP]$_2$? &  [SBJ]$_1$ is [NP]$_2$. \\
        is [SBJ]$_1$ [*\textbackslash{}NP]$_2$? &  [SBJ]$_1$ is / is not [*\textbackslash{}NP]$_2$. \\
        are [SBJ]$_1$ [*]$_2$? &  [SBJ]$_1$ are not [*]$_2$. \\
        {}[was$|$were]$_1$ [SBJ]$_2$ [*]$_3$? &  [SBJ]$_2$ [was$|$were]$_1$ [*]$_3$. \\
        {}[is$|$are$|$was$|$were]$_1$ not [SBJ]$_2$ [*]$_3$? &  [SBJ]$_2$ [is$|$are$|$was$|$were]$_1$ [*]$_3$. \\
        \midrule
        can [SBJ]$_1$ [VP]$_2$? &  [SBJ]$_1$ can [VP]$_2$. \\
        {}[MD\textbackslash{}can]$_1$ [SBJ]$_2$ [VP]$_3$? &  [SBJ]$_2$ [MD\textbackslash{}can]$_1$ / [MD\textbackslash{}can]$_1$ not [VP]$_3$. \\
        {}[MD]$_1$ not [SBJ]$_2$ [VP]$_3$? &  [SBJ]$_2$ [MD]$_1$ [VP]$_3$. \\
        \midrule
        does [SBJ]$_1$ [VP]$_2$? &  [SBJ]$_1$ does not [VP]$_2$. \\
        {}[does$|$do]$_1$ not [SBJ]$_2$ [VP]$_3$? &  [SBJ]$_2$ [VP]$_3$. \\
        {}[does$|$do]$_1$ [SBJ]$_2$ not [VP]$_3$? &  [SBJ]$_2$ [VP]$_3$. \\
        do [SBJ]$_1$ [VP]$_2$? &  [SBJ]$_1$ do / do not [VP]$_2$. \\
        did [SBJ]$_1$ [*]$_2$? &  [SBJ]$_1$ did not [*]$_2$. \\
        did not [SBJ]$_1$ [*]$_2$? &  [SBJ]$_1$ did not [*]$_2$. \\
        \bottomrule
    \end{tabularx}
    \caption{A summary of question transformation rules. Some rules have been combined into one rule expression for clarity. \textbf{(Notations)} SBJ: subject, MD: modal verb, VB: verb, VP: verb phrase, ADJ: adjective, ADV: adverb, NP: noun phrase, backslash (\textbackslash{}): exclusion. ``xxx'' and a forward slash indicate being semantically underspecified (Section \ref{sec:background}).}
    \label{tab:quest_trans_rules}
\end{table}

\end{document}